\documentclass[letterpaper,10pt]{article}
\usepackage[
    natbib=true,
    style=numeric,
    sorting=none
]{biblatex}
\usepackage{graphicx} 
\usepackage{booktabs} 
\usepackage{authblk}
\usepackage{enumitem} 
\usepackage{float} 
\date{} 
\setlist[enumerate]{leftmargin=*} 

\graphicspath{ {./} }
\usepackage{titlesec}

\titleformat{\section}
  {\normalfont\Large\bfseries}{\thesection.}{1em}{\MakeUppercase}

\usepackage[margin=1in]{geometry}

\usepackage{hyperref} 

\hypersetup{
    bookmarksopen=true, 
    bookmarksnumbered=true, 
    colorlinks=true, 
    linkcolor=black, 
    urlcolor=black, 
    citecolor=black, 
    pdfstartview={FitV}, 
    pdftitle={Title here}, 
    pdfauthor={Author here}, 
    pdfsubject={Subject here}, 
    pdfkeywords={keyword1, keyword2}, 
}

\usepackage{times}

\title{Utilizing the LightGBM Algorithm for Operator User Credit Assessment Research
}
\author[1]{Shaojie Li}
\author[2]{Xinqi Dong}
\author[3]{Danqing Ma}
\author[4]{Bo Dang}
\author[5]{Hengyi Zang}
\author[6]{Yulu Gong}
\affil[1]{Computer Technology, Huacong Qingjiao Information Technology (Beijing) Co., Ltd., Beijing, China}
\affil[2]{Management Information Systems, University of Maine at Presque Isle, Presque Isle, US}
\affil[3]{Computer Science, Stevens Intitude of technology, New Jersey, US}
\affil[4]{Computer Science, San Francisco Bay University, Fremont CA, US}
\affil[5]{Physics and Mathematics, Universitario Tecnológico Universitam, Tijuana, Mexico}
\affil[6]{Computer and Information Technology, Northern Arizona University, Flagstaff, US}
\affil[*]{Corresponding author: Shaojie Li, lishaojie@tsingj.com}

\begin{document}
\maketitle

\noindent \textbf{Abstract:} \textbf{\textit{Mobile Internet user credit assessment is an important way for communication operators to establish decisions and formulate measures, and it is also a guarantee for operators to obtain expected benefits. However, credit evaluation methods have long been monopolized by financial industries such as banks and credit. As supporters and providers of platform network technology and network resources, communication operators are also builders and maintainers of communication networks. Internet data improves the user's credit evaluation strategy.
This paper uses the massive data provided by communication operators to carry out research on the operator's user credit evaluation model based on the fusion LightGBM algorithm. First, for the massive data related to user evaluation provided by operators, key features are extracted by data preprocessing and feature engineering methods, and a multi-dimensional feature set with statistical significance is constructed; then, linear regression, decision tree, LightGBM, etc. The machine learning algorithm builds multiple basic models to find the best basic model; finally, integrates Averaging, Voting, Blending, Stacking and other integrated algorithms to refine multiple fusion models, and finally establish the most suitable fusion model for operator user evaluation.}} 

\vspace{\baselineskip} 


\section{Introduction}
\begin{flushleft}
Credit assessment originated in the United States in the early 1900s. As early as 1902, John Moody's, the founder of Moody's, first began to rate railroad securities, using empirical methods to classify the credit ratings of securities \cite{fennema2012international}. Subsequently, the credit rating method based on historical experience began to be widely used in various financial fields, and it became an important way to predict the possibility of default of large financial entities. \par

Since the last century, the basic models used in the industry and academia for personal credit assessment are mainly three types: expert scoring model (ES-Model, Expert Scoring Model) \cite{biccer2010bayesian}, statistical model (S-Model, Statistical Model)\cite{konishi1999statistical} and artificial intelligence model (AI-Model, Artificial Intelligence Model)\cite{punniyamoorthy2016identification}. \par

Due to the advantages of data processing efficiency and accuracy, the credit assessment method based on machine learning has played an increasingly important role in the credit assessment industry. The credit assessment method has shifted from the traditional experience-driven to data-driven. In practical applications, due to the continuous growth of data volume and application requirements\cite{hofmann2017big}, a single machine learning method has been unable to meet the growing requirements of engineering problems. The fusion algorithm based on multiple machine learning knowledge and feature engineering has become a new research focus. These fusion models that "learn from the strengths of others" provide new ideas for solving credit assessment problems. \par

In terms of personal user credit assessment, communication operators have unique advantages. The ubiquitous mobile payment helps operators to explore the flow of personal users' wealth, the widely constructed telecommunications base stations help operators to grasp the traffic trends of personal users, and the personal information continuously pouring in from the mobile Internet makes it possible for operators to observe personal users in almost every aspect. Compared with banks that only grasp the financial behavior of personal users, operators have an advantage in data volume\cite{chen2016financial}, but still face difficulties in the rational use of data. \par

In the pursuit of operator user credit assessment research, employing the LightGBM algorithm and integrating the latest research findings and methodologies is essential. The studies by Zhu et al. emphasize the enhancement of credit prediction models by incorporating various ensemble methods to improve accuracy and performance \cite{zhu2024ensemble}. Additionally, another study by Zhu et al. presents a method that synergizes neural networks with the Synthetic Minority Over-sampling Technique (SMOTE), effectively boosting the detection capabilities for credit card fraud \cite{zhu2024enhancing}. These innovative approaches offer valuable insights and applicable methodologies for operator user credit assessment. \par

In actual use, due to different scenarios, data, etc., a single model cannot achieve ideal results in all scenarios. Therefore, the model fusion theory is proposed, which aims to integrate the advantages of different models, construct an ensemble learning method (Ensemble Learning)\cite{sagi2018ensemble}, and form a lower variance, smaller deviation, and better performance ensemble model (Ensemble Model).
\end{flushleft}

\section{Related Work}

\subsection{Random forest}
\begin{flushleft}
Random forest is an integrated algorithm that uses a collection of CART decision trees for classification and regression\cite{breiman2001random}. Due to the introduction of randomly generated samples, the overfitting situation of the decision tree algorithm can be greatly improved. Random forest uses the Bagging homogeneous integration method to reduce variance while maintaining low bias.
\end{flushleft}

\subsection{GBDT algorithm}
\begin{flushleft}
It is not difficult to see that the GBDT algorithm is an ensemble algorithm based on trees similar to random forest \cite{si2017gradient}. The difference is that the Boosting integration is used to train a group of decision trees in sequence. Each subsequent tree will reduce the error of the previous tree by using the residual of the previous model to fit the next model. Since GBDT is trained sequentially, it is generally believed that it is slower and less scalable than random forest, which can train multiple trees in parallel. However, compared with random forest, GBDT usually uses shallower trees, which means that GBDT can train faster. Increasing the number of trees in GBDT will increase the chance of overfitting (GBDT reduces deviation by using more trees), while increasing the number of trees in random forest will reduce the chance of overfitting (random forest uses more trees to reduce variance).
\end{flushleft}

\subsection{XGBoost algorithm}
\begin{flushleft}
XGBoost (eXtreme Gradient Boosting) is one of the best GBDT implementations available today\cite{chen2016xgboost}. Its introduced parallel tree enhancement feature makes it much faster than other tree-based ensemble algorithms in terms of establishment speed. In 2015, 17 of the 29 winning solutions of Kaggle used XGBoost, and the top 10 solutions of the 2015 KDD Cup used XGBoost.

XGBoost is designed using the general principle of gradient boosting, which combines weak learners into strong learners. Although GBDT is constructed sequentially - learning slowly from the data to improve its predictions in subsequent iterations, the feature-grained parallel construction is implemented on XGBoost. Because decision tree learning takes a lot of time in the best split of features, XGBoost adopts the method of data pre-sorting (Pre-Sorted Method), saves the sorted data information as block units, and supports the repeated use of the subsequent iteration process. The existence of this block unit makes it possible to perform multi-threaded calculations in parallel CPUs for feature gain calculations.

XGBoost is also cache-aware, can reduce overfitting by controlling model complexity and built-in regularization, effectively handle sparse data, and can use disk space (rather than just memory) for large datasets to support out-of-core computing, thereby maximizing system computing capabilities and producing better prediction performance.
\end{flushleft}

\subsection{LightGBM algorithm}
\begin{flushleft}
LightGBM is a lightweight gradient boosting algorithm similar to XGBoost. It was released on October 17, 2016 as part of Microsoft's Distributed Machine Learning Toolkit (DMTK)\cite{ke2017lightgbm}. It is designed to be fast and distributed, so it has faster training speed and lower memory usage, and supports GPU and parallel learning at the same time. The ability to process large datasets.

LightGBM has been proven to be faster and more accurate than XGBoost in benchmarks and experiments on multiple public datasets. LightGBM has several advantages over XGBoost. It uses histograms to store continuous features into discrete bins (discrete binning method), which provides LightGBM with several performance advantages over XGBoost (which uses a pre-sort-based algorithm for tree learning by default), such as reducing memory usage, reducing the cost of calculating the gain of each split, and reducing the communication cost of parallel learning.

LightGBM calculates the histogram of a node by performing histogram subtraction on its sibling nodes and parent nodes, which allows the node histograms to be reused (only one node needs to be built for each split), resulting in additional performance improvements. Existing benchmarks show that LightGBM is 11 to 15 times faster than XGBoost on some tasks. In addition, the LightGBM algorithm uses a leaf-wise growth (Leaf Wise) strategy, which usually converges faster and achieves lower losses, and is generally more accurate than the layer-wise growth of XGBoost.

Since it is an engineering implementation of GBDT, LightGBM also has a high degree of freedom in parameter settings, which will help to establish a machine learning model with better performance using LightGBM. The following parameters can usually be set:

\begin{table}[H]
\centering 
\caption{ LightGBM parameter table} 
\label{tab:model_parameters}
\begin{tabular}{@{}lp{10cm}@{}} 
\toprule 
\textbf{Parameter} & \textbf{Description} \\
\midrule 
Parameter          & Description \\
Max Depth          & Set this parameter to prevent the tree from growing too deep, thereby reducing the risk of overfitting. \\
Num Leaves         & Controls the complexity of the tree model. Setting it to a larger value can improve accuracy but may increase the risk of overfitting. \\
Min Data in Leaf   & Setting this parameter to a larger value can prevent the tree from growing too deep. \\
Max Bin            & Controls the number of discrete bins. Smaller values can control overfitting and speed up training, while larger values can improve accuracy. \\
Feature Fraction   & Enables random feature subsampling. Reasonably setting this can speed up training and prevent overfitting. \\
Bagging Fraction   & Specifies the fraction of data to be used for each iteration. Reasonably setting this can speed up training and prevent overfitting. \\
Num Iteration      & Sets the number of boosting iterations. Setting this parameter affects the training speed. \\
Objective          & Set this parameter to specify the type of task the model attempts to perform. \\
\bottomrule 
\end{tabular}
\end{table}

\end{flushleft}

\section{Data Processing and Feature Selection}
\begin{flushleft}
The operator user credit assessment analyzes user behavior data and historical records on the operator side to assess user credit status. Feature engineering is a key link in this process, which aims to mine and extract valuable features so as to more accurately portray user credit status. \par

Before embarking on model building, it is essential to have a deep understanding of the basic attributes and characteristics of the required data, because this lays a solid foundation for subsequent data preprocessing, feature engineering and modeling work. By analyzing the distribution, outliers, missing conditions, and correlation of data, we can formulate more efficient data processing strategies. This helps us accurately identify and extract key features that have a significant impact on model performance, and thus build a more reliable and better performing credit assessment model. The operator user data fields are shown in Table 2. \par 

\begin{table}[H]
\centering 
\caption{Feature field description} 
\begin{tabular}{@{}lllllll@{}}
\toprule 
\textbf{Number} & \textbf{Feature} & \textbf{Feature Description} \\
\midrule 
1 & id & User ID \\
2 & age & User Age \\
3 & net\_age\_till\_now & User's Network Age (months) \\
4 & top\_up\_month\_diff & Months Since Last Top-Up \\
5 & top\_up\_amount & Amount of Last Top-Up (CNY) \\
6 & recent\_6month\_avg\_use & Avg Spending on Calls (Last 6 Months, CNY) \\
7 & total\_account\_fee & Total Bill for the Current Month (CNY) \\
8 & curr\_month\_balance & Current Month Account Balance (CNY) \\
9 & connect\_num & Number of Contacts This Month \\
10 & recent\_3month\_shopping\_count & Avg Monthly Shopping Visits (Last 3 Months) \\
11 & online\_shopping\_count & Online Shopping App Uses This Month \\
12 & express\_count & Express Delivery App Uses This Month \\
13 & finance\_app\_count & Financial Management App Uses This Month \\
14 & video\_app\_count & Video Streaming App Uses This Month \\
15 & flight\_count & Air Travel App Uses This Month \\
16 & train\_count & Train App Uses This Month \\
17 & tour\_app\_count & Travel Info App Uses This Month \\
18 & cost\_sensitivity & Sensitivity to Phone Bill Costs \\
19 & score & Credit Score (Prediction Target) \\
20 & true\_name\_flag & Passed Real-Name Verification \\
21 & uni\_student\_flag & University Student Status \\
22 & blk\_list\_flag & Blacklist Status \\
23 & 4g\_unhealth\_flag & 4G Unhealthy Customer Status \\
24 & curr\_overdue\_flag & Current Overdue Payment Status \\
25 & freq\_shopping\_flag & Frequent Shopping Mall Visitor \\
26 & wanda\_flag &Visited Fuzhou Cangshan Wanda This Month \\
27 & sam\_flag & Visited Fuzhou Sam's Club This Month \\
28 & movie\_flag & Watched a Movie This Month \\
29 & tour\_flag & Visited Tourist Attractions This Month \\
30 & sport\_flag & Consumed at Sports Venues This Month \\
\bottomrule 
\end{tabular}
\end{table}

Feature selection is the process of selecting a subset of features from the original feature set. An ideal feature subset should be as low in dimensionality as possible and as statistically significant as possible. In order to improve the statistical significance of the feature subset and reduce the dimensionality of the feature set, the feature set was artificially divided into four feature subsets, which are also the four dimensions of the operator's user portrait in this paper. \par

When performing feature selection, it is not necessary to perform feature segmentation manually. Using methods such as Bagging random sampling may achieve better results when establishing the model, but using random methods will ignore the aspects that have been marked in the data, and the basic models established will lose statistical significance. In order to better observe the correlation between different types of features and credit scores, manual segmentation is selected here.

\begin{table}[H]
\centering 
\caption{Feature subsets} 
\begin{tabular}{@{}lp{10cm}@{}} 
\toprule 
\textbf{Number} & \textbf{Feature subset} \\
\midrule 
1	&Consumer Capacity	\\
2	&Location Trajectory	 \\
3	&Application Behavior Preference	 \\
4	&Other \\
\bottomrule 
\end{tabular}
\end{table}

\end{flushleft}

\section{Results and Analysis}
\begin{flushleft}
The following will compare the simulation results of the machine learning models . The main evaluation indicators include MAE (mean absolute error), MAPE (mean absolute percentage error), MSE (mean squared error), RMSE (root mean squared error) and R2 (accuracy).
\subsection{Consumption Capacity} 
The consumer capacity dataset is mainly used to measure the user's related characteristics in the mobile communication business. The basic models of dataset 1 are constructed using linear regression, decision tree, random forest and LightGBM algorithms, respectively. The experimental results are shown in Table 4 and Figure 1.

\begin{table}[H]
\centering
\caption{Comparison of models of consumer capacity}
\label{tab:consumer_capacity}
\begin{tabular}{@{}lllllll@{}}
\toprule
Dataset & Method & MAE & MAPE & MSE & RMSE & \( R^2 \) \\
\midrule
Consumer Capacity & Linear Regression (LR) & 28.0529 & 0.4692 & 1306.8418 & 36.1503 & 0.2793 \\
Consumer Capacity & Decision Tree (DT) & 24.0616 & 0.4013 & 1000.3104 & 31.6277 & 0.4483 \\
Consumer Capacity & Random Forest (RF) & 23.7165 & 0.3957 & 953.7934 & 30.8835 & 0.4740 \\
Consumer Capacity & LightGBM (LGB) & 22.1867 & 0.3701 & 935.7114 & 30.5894 & 0.6534 \\
\bottomrule
\end{tabular}
\end{table}

\begin{figure}[H]
    \centering
    \includegraphics[scale=0.6]{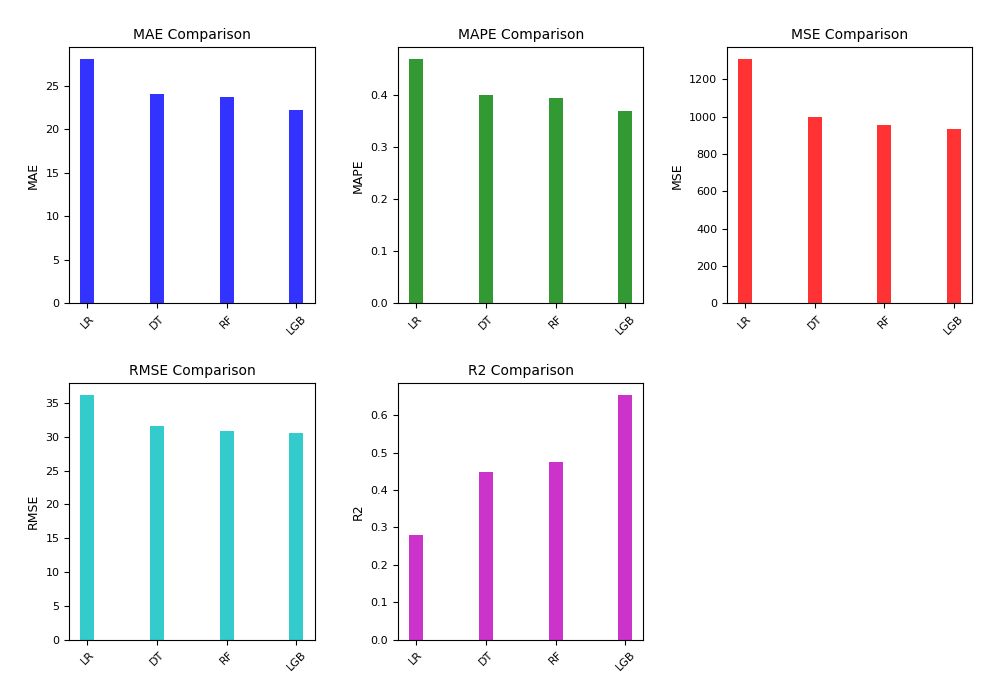}
    \caption{Comparison of basic models of consumer capacity}
\end{figure}

In the model prediction of consumer capacity, it can be found that LightGBM's indicators are far better than the LR, DT, and RF basic models, proving that it does have good accuracy.

\subsection{Location Trajectory}
The location trajectory dataset is mainly used to measure the user's daily activity location trajectory related characteristics. The basic models of dataset 2 are constructed using linear regression, decision tree, random forest and LightGBM algorithms, respectively. The experimental results are shown in Table 5 and Figure 2.

\begin{table}[H]
\centering
\caption{Comparison of models of location trajectory}
\label{tab:location_trajectory}
\begin{tabular}{@{}lllllll@{}}
\toprule
Dataset & Method & MAE & MAPE & MSE & RMSE & \( R^2 \) \\
\midrule
Location Trajectory & Linear Regression (LR) & 31.6287 & 0.5296 & 1619.6993 & 40.2455 & 0.1067 \\
Location Trajectory & Decision Tree (DT) & 31.0083 & 0.5191 & 1570.8188 & 39.6336 & 0.1337 \\
Location Trajectory & Random Forest (RF) & 30.9696 & 0.5186 & 1565.0511 & 39.5607 & 0.1369 \\
Location Trajectory & LightGBM (LGB) & 30.7300 & 0.5144 & 1547.7564 & 39.3415 & 0.1464 \\
\bottomrule
\end{tabular}
\end{table}

\begin{figure}[H]
    \centering
    \includegraphics[scale=0.6]{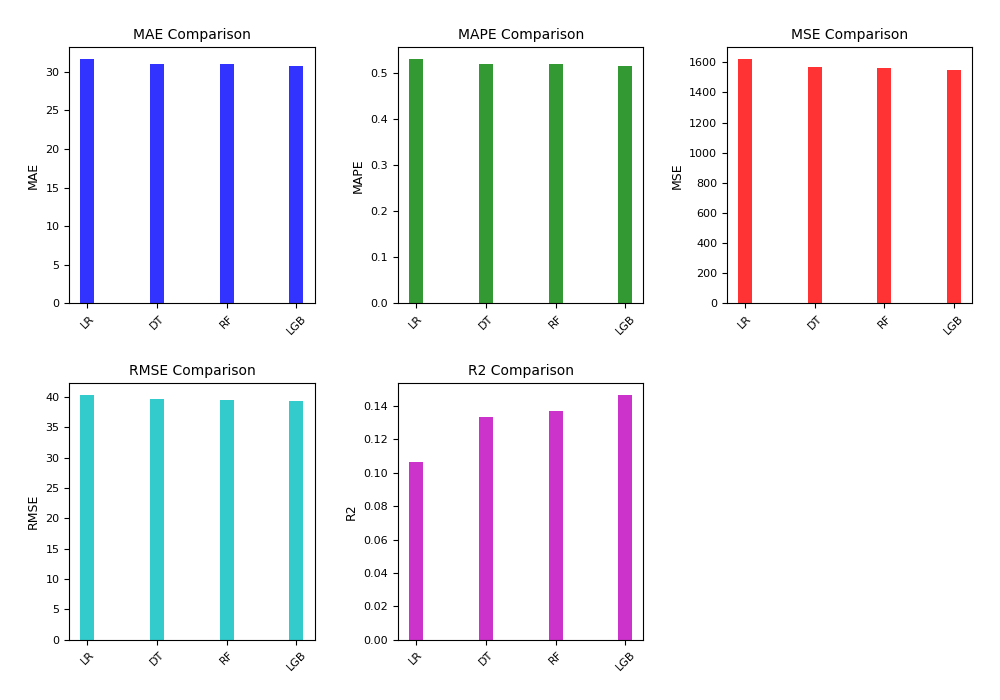}
    \caption{Comparison of models of location trajectory}
\end{figure}

In the model prediction of location trajectory, the performance of the four algorithms is not satisfactory, which may be due to the low correlation between location trajectory and user credit, but LightGBM's indicators are still slightly better than the LR, DT, and RF basic models.

\subsection{Application Behavior Preference}
Application preference is mainly used to reflect the user's application usage, and the basic models of dataset 3 are constructed using linear regression, decision tree, random forest and LightGBM algorithms, respectively. The experimental results are shown in Table 6 and Figure 3.

\begin{table}[H]
\centering
\caption{Comparison of models of application behavior preference}
\label{tab:application_preference}
\begin{tabular}{@{}lllllll@{}}
\toprule
Dataset & Method & {MAE} & {MAPE} & {MSE} & {RMSE} & {$R^2$} \\
\midrule
Application Behavior Preference & Linear Regression (LR) & 33.3207 & 0.5579 & 1776.5497 & 42.1491 & 0.0202 \\
Application Behavior Preference & Decision Tree (DT) & 28.9453 & 0.4833 & 1400.0899 & 37.4178 & 0.2279 \\
Application Behavior Preference & Random Forest (RF) & 28.6198 & 0.4779 & 1353.8066 & 36.7941 & 0.2534 \\
Application Behavior Preference& LightGBM (LGB) & 23.1702 & 0.3881 & 973.7953 & 31.2057 & 0.4630 \\
\bottomrule
\end{tabular}
\end{table}

\begin{figure}[H]
    \centering
    \includegraphics[scale=0.6]{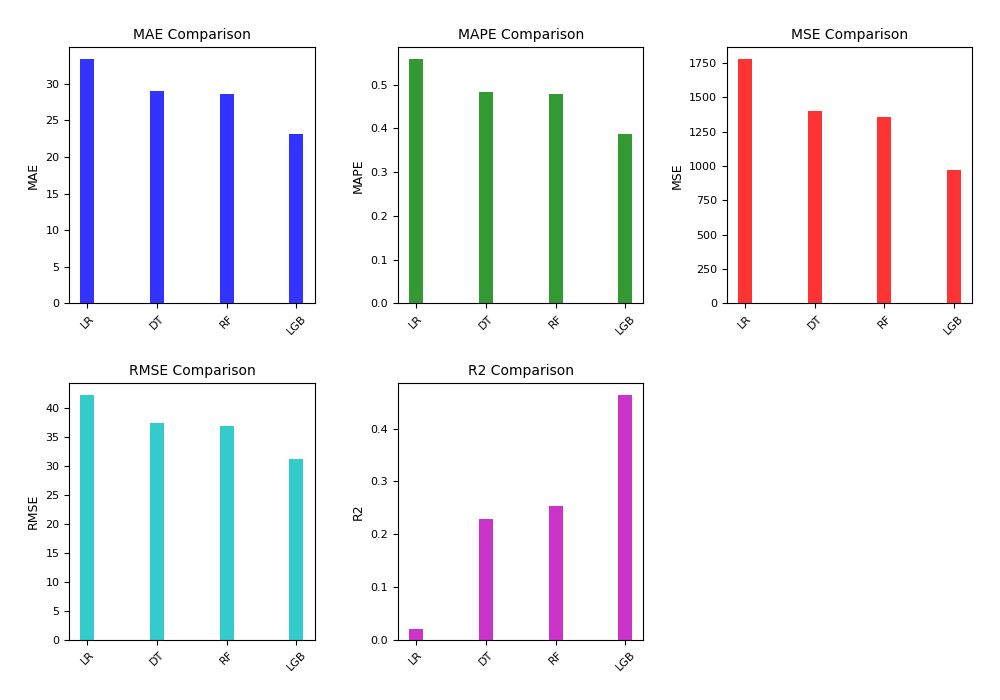}
    \caption{Comparison of models of application behavior preference}
\end{figure}

In the model prediction of application behavior preference, it can be found that the effect of the basic model established by linear regression is very poor, indicating that the nonlinearity of this dataset is very high, and LightGBM's indicators are far better than the LR, DT, and RF basic models, proving Its processing performance on nonlinear data.

\subsection{Other}
The other model contains a large number of features built by feature construction and feature extraction processes, which are highly correlated with each other. The basic models of dataset 4 - Other are constructed using linear regression, decision tree, random forest and LightGBM algorithms, respectively. The experimental results are shown in Table 7 and Figure 4.

\begin{table}[H]
\centering
\caption{Comparison of basic models of others}
\label{tab:others}
\begin{tabular}{@{}lllllll@{}}
\toprule
Dataset & Method & MAE & MAPE & MSE & RMSE & \( R^2 \) \\
\midrule
Other & Linear Regression (LR) & 24.6979 & 0.4126 & 1015.5399 & 31.8675 & 0.4399 \\
Other & Decision Tree (DT) & 23.7093 & 0.3926 & 965.4682 & 31.0720 & 0.4671 \\
Other & Random Forest (RF) & 22.2056 & 0.3786 & 942.3304 & 30.6974 & 0.6436 \\
Other & LightGBM (LGB) & 17.5637 & 0.2911 & 539.5605 & 23.2284 & 0.7024 \\
\bottomrule
\end{tabular}
\end{table}

\begin{figure}[H]
    \centering
    \includegraphics[scale=0.6]{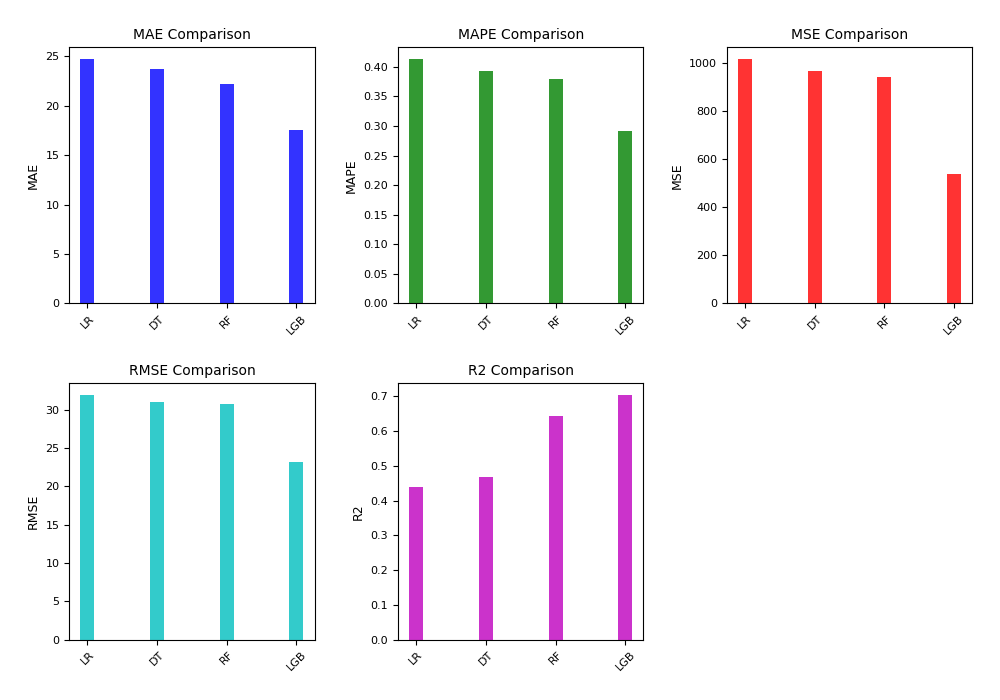}
    \caption{Comparison of basic models of others}
\end{figure}

In the model prediction of other types of data, it can be found that LightGBM still has advantages in various indicators, reflecting the effect of the mutual exclusion feature bundling algorithm in the LightGBM algorithm on highly correlated features.

This chapter compares machine learning algorithms such as LightGBM, random forest, decision tree, and linear regression to construct credit assessment basic models, and evaluates the performance of basic models of multiple data sets, and finally selects the best basic model for subsequent chapters. Fusion model establishment.
\end{flushleft}

\subsection{Constructing an fusion LightGBM model}
\begin{flushleft}
The fusion model is a type of ensemble learning model that integrates multiple base models. By further learning from the strengths of different base models, it achieves superior performance. In the work carried out in the previous section, credit evaluation models have been constructed that target aspects such as consumer capacity, location trajectories, and application behavior preferences. This series of models can serve as the basis for establishing a fusion model.

The establishment of the fusion model begins by splitting the dataset, after feature engineering, into four subsets: consumer capability, location trajectory, application behavior preference, and others. LightGBM is then applied to each of these subsets to learn from them, and we successfully obtain four corresponding base models for user credit evaluation. To achieve a more refined credit assessment fusion model, we introduced three representative ensemble algorithms as secondary learners: Voting, Blending, and Stacking. By applying these algorithms, we train meta-models (secondary learners) based on the four base models, which then form the final credit evaluation model. Lastly, we conduct a comparative analysis of their performance. This will further enrich our research results in the field of credit assessment and provide operators with a more comprehensive and precise basis for credit evaluation. The experimental results are shown in Table 8.

\begin{table}[H]
\centering
\caption{Comparison of ensemble model of full dataset}
\label{tab:others}
\begin{tabular}{@{}lllllll@{}}
\toprule
Dataset & Method & {MAE} & {MAPE} & {MSE} & {RMSE} & {$R^2$} \\
\midrule
Dataset 4 Only & LightGBM & 17.5637 & 0.2911 & 539.5605 & 23.2284 & 0.7024 \\
Full Dataset & LightGBM & 17.4223 & 0.2889 & 522.6124 & 22.8607 & 0.7118 \\
Full Dataset & LightGBM + Voting & 17.0166 & 0.2828 & 514.3879 & 22.6801 & 0.7163 \\
Full Dataset & LightGBM + Blending & 14.3725 & 0.2381 & 340.6276 & 18.4561 & 0.8157 \\
Full Dataset & LightGBM + Stacking & 13.1022 & 0.2166 & 311.8294 & 17.6587 & 0.8280 \\
\bottomrule
\end{tabular}
\end{table}

The experimental results show that the modeling effects of using ensemble learning techniques are generally better than those without ensemble learning. Even the simplest LightGBM-Voting ensemble algorithm outperforms the global LightGBM model. Among the ensemble learning effects, Voting is significantly behind Stacking and Blending, proving that establishing a secondary learner and implementing weight updates is a crucial aspect of the ensemble; although it takes more modeling time, Stacking is slightly superior to Blending in all metrics. Therefore, it is considered that the LightGBM-Stacking ensemble method is the most advantageous in credit evaluation models.
\end{flushleft}

\section{Conclusion}
\subsection{Summary}
The construction of the credit assessment model of the operator's users is an important direction for the informatization and digital transformation of the operator, and it is also an important way for the operator to ensure the expected benefits. This paper uses feature engineering technology, multiple machine learning technologies and multiple ensemble learning technologies to build a mixed model for operator user credit assessment, and conducted multiple comparisons, which are mainly divided into the following three aspects:
\begin{enumerate}[label=(\arabic*)]
\item  Application research of feature engineering: In order to solve the problem of feature extraction of desensitized data of personal users' mobile Internet usage, this paper uses feature engineering methods to establish features of available parts of data, and uses database classification methods to establish multiple subsets of the original data set, not only retains the statistical significance of the data, but also improves the usability of the data, and has significantly improved the establishment time and accuracy compared with the traditional scheme.
\item  Research on supervised learning modeling: At present, most of the domestic analysis and research on credit assessment still adopt the first-generation supervised learning model that is not combined with ensemble learning. In order to solve the problem of massive data feature modeling, this paper conducts multiple modeling on different types of features and selects the best ones, which has better accuracy, and the model is more in line with the actual situation and has better reference.
\item Different Ensemble Algorithm Research: Currently, most traditional applications that do not involve the segmentation and integration of multi-dimensional datasets adopt only one type of ensemble method or do not use ensemble methods at all for modeling. This approach may lead to performance overlaps in models when dealing with complex data characteristics. To address the challenge of integrating massive data features, this paper builds on the introduction of the Boosting ensemble method within the LightGBM algorithm and further implements Stacking/Blending integration from model fusion theory to establish the final model. By comparing with the Averaging/Voting ensemble and a global LightGBM algorithm that only utilizes Boosting ensemble, it is demonstrated that the established fusion model exhibits superior performance.
\end{enumerate}
\newpage

\printbibliography 

@inproceedings{chen2016xgboost,
  title={Xgboost: A scalable tree boosting system},
  author={Chen, Tianqi and Guestrin, Carlos},
  booktitle={Proceedings of the 22nd acm sigkdd international conference on knowledge discovery and data mining},
  pages={785--794},
  year={2016}
}

@book{fennema2012international,
  title={International networks of banks and industry},
  author={Fennema, Meindert},
  volume={2},
  year={2012},
  publisher={Springer Science \& Business Media}
}

@inproceedings{biccer2010bayesian,
  title={Bayesian credit scoring model with integration of expert knowledge and customer data},
  author={Bi{\c{c}}er, I{\c{s}}{\i}k and Sevis, Deniz and Bilgi{\c{c}}, Taner},
  booktitle={International Conference 24th ini EURO Conference “Continuous Optimization and Information Technologies in the FFinancial Sector”(MEC EurOPT 2010)},
  pages={324--329},
  year={2010}
}

@incollection{konishi1999statistical,
  title={Statistical model evaluation and information criteria},
  author={Konishi, Sadanori},
  booktitle={Multivariate Analysis, Design of Experiments, and Survey Sampling},
  pages={393--424},
  year={1999},
  publisher={CRC Press}
}

@article{punniyamoorthy2016identification,
  title={Identification of a standard AI based technique for credit risk analysis},
  author={Punniyamoorthy, Murugesan and Sridevi, P},
  journal={Benchmarking: An International Journal},
  volume={23},
  number={5},
  pages={1381--1390},
  year={2016},
  publisher={Emerald Group Publishing Limited}
}

@article{hofmann2017big,
  title={Big data and supply chain decisions: the impact of volume, variety and velocity properties on the bullwhip effect},
  author={Hofmann, Erik},
  journal={International Journal of Production Research},
  volume={55},
  number={17},
  pages={5108--5126},
  year={2017},
  publisher={Taylor \& Francis}
}

@inproceedings{si2017gradient,
  title={Gradient boosted decision trees for high dimensional sparse output},
  author={Si, Si and Zhang, Huan and Keerthi, S Sathiya and Mahajan, Dhruv and Dhillon, Inderjit S and Hsieh, Cho-Jui},
  booktitle={International conference on machine learning},
  pages={3182--3190},
  year={2017},
  organization={PMLR}
}

@article{ke2017lightgbm,
  title={Lightgbm: A highly efficient gradient boosting decision tree},
  author={Ke, Guolin and Meng, Qi and Finley, Thomas and Wang, Taifeng and Chen, Wei and Ma, Weidong and Ye, Qiwei and Liu, Tie-Yan},
  journal={Advances in neural information processing systems},
  volume={30},
  year={2017}
}

@article{sagi2018ensemble,
  title={Ensemble learning: A survey},
  author={Sagi, Omer and Rokach, Lior},
  journal={Wiley Interdisciplinary Reviews: Data Mining and Knowledge Discovery},
  volume={8},
  number={4},
  pages={e1249},
  year={2018},
  publisher={Wiley Online Library}
}

@article{chen2016financial,
  title={Financial credit risk assessment: a recent review},
  author={Chen, Ning and Ribeiro, Bernardete and Chen, An},
  journal={Artificial Intelligence Review},
  volume={45},
  pages={1--23},
  year={2016},
  publisher={Springer}
}

@article{breiman2001random,
  title={Random forests},
  author={Breiman, Leo},
  journal={Machine learning},
  volume={45},
  pages={5--32},
  year={2001},
  publisher={Springer}
}

@article{zhu2024ensemble,
  title={Ensemble Methodology: Innovations in Credit Default Prediction Using LightGBM, XGBoost, and LocalEnsemble},
  author={Zhu, Mengran and Zhang, Ye and Gong, Yulu and Xing, Kaijuan and Yan, Xu and Song, Jintong},
  journal={arXiv preprint arXiv:2402.17979},
  year={2024}
}

@article{zhu2024enhancing,
  title={Enhancing Credit Card Fraud Detection: A Neural Network and SMOTE Integrated Approach},
  author={Zhu, Mengran and Zhang, Ye and Gong, Yulu and Xu, Changxin and Xiang, Yafei},
  journal={Journal of Theory and Practice of Engineering Science},
  volume={4},
  number={02},
  pages={23--30},
  year={2024}
}

\end{document}